\pdfoutput=1

\documentclass[11pt]{article}
\usepackage[final]{acl}
\usepackage{graphicx}

\usepackage{times}
\usepackage{latexsym}

\usepackage[T1]{fontenc}
\usepackage[utf8]{inputenc}

\usepackage{microtype}

\usepackage{amsmath}
\title{A Classification System Approach in Predicting Chinese Censorship \\ \vspace*{3pt} \large \textsc{New York University} \vspace*{-20pt}}

\author{Tianchu Ze \\
  \texttt{tz2018@nyu.edu} \And
  Yushen Hu  \\
  \texttt{yh3856@nyu.edu} \And
  Matt Prodani\\
  \texttt{mattp@nyu.edu}}

\begin{document}
\maketitle
\vspace*{-40pt}
\begin{abstract}
This paper is dedicated to using a classifier to predict whether a Weibo post would be censored under the Chinese internet. Through randomized sampling from \citeauthor{Fu2021} and Chinese tokenizing strategies, we constructed a cleaned Chinese phrase dataset with binary censorship markings. Utilizing various probability-based information retrieval methods on the data, we were able to derive 4 logistic regression models for classification. Furthermore, we experimented with pre-trained transformers to perform similar classification tasks. After evaluating both the macro-F1 and ROC-AUC metrics, we concluded that the Fined-Tuned BERT model exceeds other strategies in performance.
\end{abstract}

\section{Introduction}

Introduced around the start of the 21st Century,  China’s national firewall had been acting as a means of censorship for foreign websites and sensitive information. Without any specific political message or inclination, our project is driven by a recent upgrade of “security measures” for the firewall by Beijing. In other words, more sensitive keywords are incorporated into the censorship measurement. Similar updates had been observed in history, most recently in 2012. Briefly, that historical tightening of security was due to a shift of power within the Chinese government and a rumored coup d'etat. The event happening from November through December of 2022 is not a political discourse, but more of a rise of tension within the general public due to China’s zero-tolerance restriction enforcement across the country to prevent COVID-19. 

Again, this paper is not intended to make a political voice, but to explore the feasibility of reverse engineering a censorship label system using NLP techniques and ML classification libraries. Therefore, the paper will look into various ways of modeling strategies to find the highest-scored system. This project will incorporate data sources from various data sources on different dimensions of the Chinese internet, but mostly from the blog website, Weibo. The platform is considered the Twitter of China and had been referenced in multiple prior pieces of research. We do acknowledge that different access times of the platform may or most likely will contain different subsets of censored keywords or queries. But, our goal is to identify such supersets that will be banned undoubtedly. For example, information in regards to the June 4th protests against Beijing. Additionally, data sources referenced within this paper mostly come from the western world and large discrepancies will persist. After all, our priority of the project is not to fully reconstruct the censorship library, but more on exploring if such a censorship annotator is feasible. 
\subsection{Workflow}
\begin{figure}[!htb]
\includegraphics[width=220px]{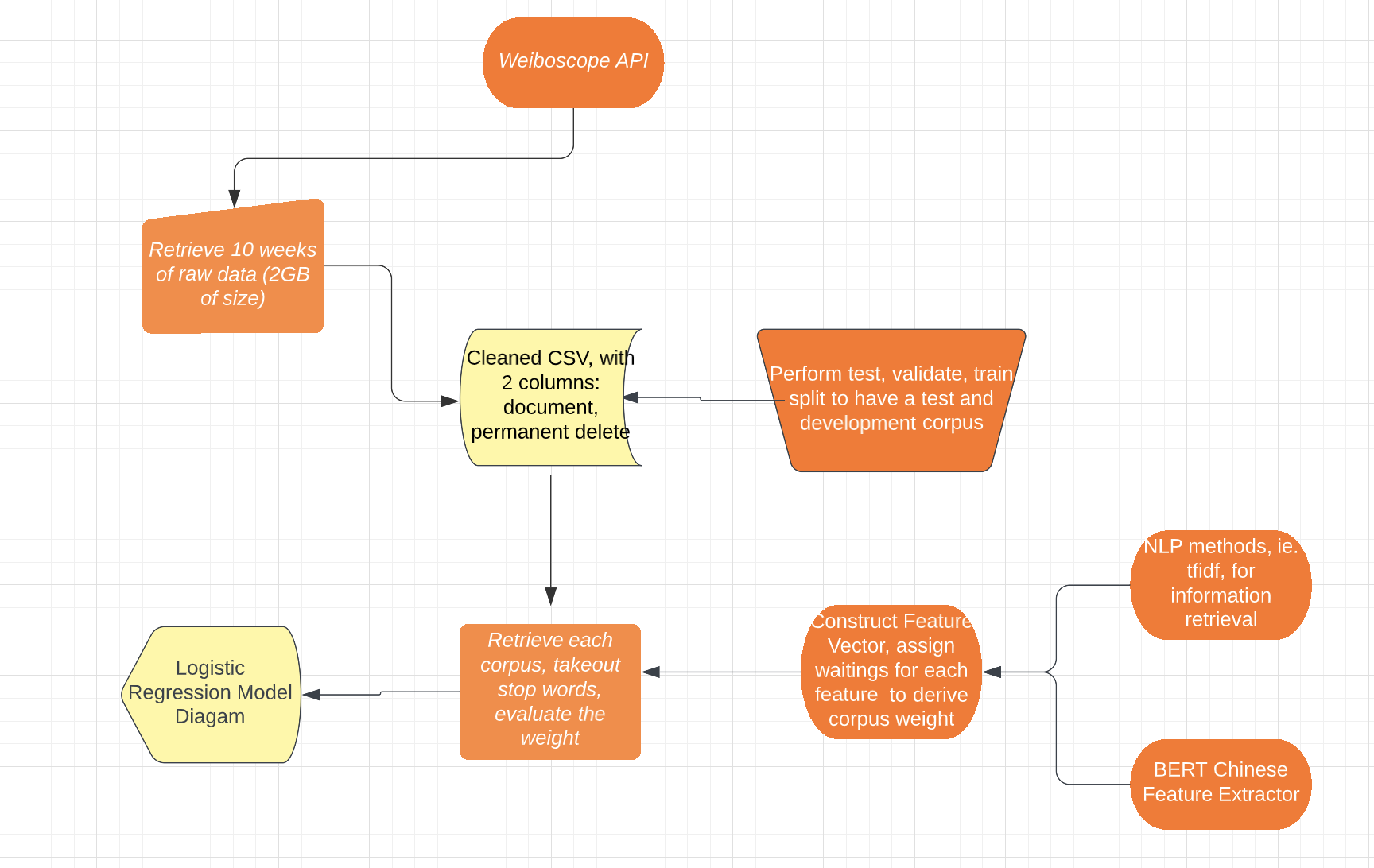}
\label{Fig. 1} 
\caption{Team Workflow Diagram}
\end{figure}
Our paper will be employing the \citeauthor{Fu2021} API, which will be detailed in the following sections on data profiling. After splitting the test and development corpus, we will work with the development corpus and leave the testing one to the side. Before we perform any NLP methods, taking out the stop words and tokenizing each meaningful Chinese phrase is crucial. Due to the nature of the Chinese language, each character may mean something by itself or combined with its neighbors; there is no clear law in which the characters are combined. Therefore, we have employed a proven NLP library, \citeauthor{Jiebam}, to perform such tokenization for us . After the split, 4 variations of information retriever inspired by a TFIDF approach from previous research will be applied to the training data from the development corpus to find feature vectors for each row \cite{w2}. Using the resulting features, a regression model will be constructed for all 4 retrievers to be evaluated. Furthermore, a model employing Bert  will be considered as well for evaluation \cite{bertpaper}.

\section{Related Works}
\citeauthor{w1} provides an overview of censorship practices within the microblogs of Weibo. Specifically, it is a term extraction project, which showcased a list of censored topics or keywords for 2012, a period of tightening in internet censorship. This list mostly consists of political references to CCP scandals or the rumored coup d'etat, both strictly censored topics. But in terms of application to our research, this list was derived from \citeauthor{Fu2021}. It is a data collection system built on Sina’s open API, that could track the timeline of thousands of users’ blogs and identify their removal status by repeated accessing. With over 111 million blogs collected in the year 2012, the API may serve as a great sample for the modeling of our classifier system.

\citeauthor{w2} serves as a basis for our research, as it suggests and investigates the hypothesis of the existence of a Chinese topic surveillance list. Additionally, it also employs \citeauthor{Fu2021} along with the use of TF-IDF in topic and feature extraction. The paper concludes that the platform supports both explicit, implicit, and camouflaged filtering based on the content of each post. However, the paper mentioned their difficulties in the tokenization of Chinese text, as well as stopped at the stage of finding the list of term frequency for each word. Therefore, inspired by the paper, we continued after the TF-IDF stage and used these features to construct a logistic regression model. Lastly, the paper also mentioned how the user may play a role in censoring. If a user had been posting censored content, that account’s posts may be auto-deleted. This may be something to be noted during further research.

\citeauthor{ahmed-kumar-m-2021-classification} serves as another basis for our research, as it employs pre-trained transformer models in XLNet, an auto-regressive model, to perform feature extraction as well as modeling. However, in this paper, our focus is on the evaluation scores. As their dataset is also implemented with \citeauthor{Fu2021} and another modern library, their results may be used for comparison with others. Specifically, the authors also have implemented BERT as a part of their research for comparisons with XLNet. The highest F-1 score of their system on the validation data is 0.634, and that is our initial goal. Furthermore, the paper also influenced performing binary classification, as the result of our data processing is in a binary form. However, a classical logistic regression is chosen compared to XLNet’s regressor due to our implementation of a TF-IDF-like feature extractor along with Bert.

\citeauthor{10.1145/3442188.3445916} uses a similar approach to what we are planning to use, it uses word embedding to calculate the similarity between Wikipedia and Baidu Baike, a Chinese version of Wikipedia. It identifies several target words that are mainly democratic political terms and propaganda. By calculating the cosine similarities, this paper deduces the differences in p-value and effect size of these words between Wikipedia and Baidu Baike. However, regarding our research goal, its main focus is on online encyclopedias, which are pre-written texts and lack user interactions. For our project, we are planning to apply a similar model to social media in China, in which the model provided by this paper will be helpful.

In \citeauthor{33531}, rather than research based on the gathering of data, the author’s investigation of the censorship coverage on different Chinese sites and magnitude may be a reference for our weighting of each site’s result during evaluation. Another interesting field that the authors employed within the paper is categorizing pornography. In fact, it is shown that there is a higher percentage of pattern matches with pornography keywords than political messages. We may also consider taking account of keywords relating to such. 

\section{Dataset Preprocessing}

\subsection{Preprocessing}
As seen in figure 2, the \citeauthor{Fu2021} data consists of various columns, ranging from the user to the timestamp. Apart from the actual text by the user, we also need to find another column that will determine if the text is censored or not. The permission-denied column is used to determine such status, if a 1 appeared, that means the official Weibo open API returned such status upon request. As mentioned above in related works, \citeauthor{Fu2021} API is built on the official API and tracks the timeline of each post of each tracked user during the period of 2012. Unfortunately, this long period may result in over 111 million rows of data, which we were able to decrease in size significantly with randomization under a limited sample size.
\begin{figure}[!htb]
\includegraphics[width=220px]{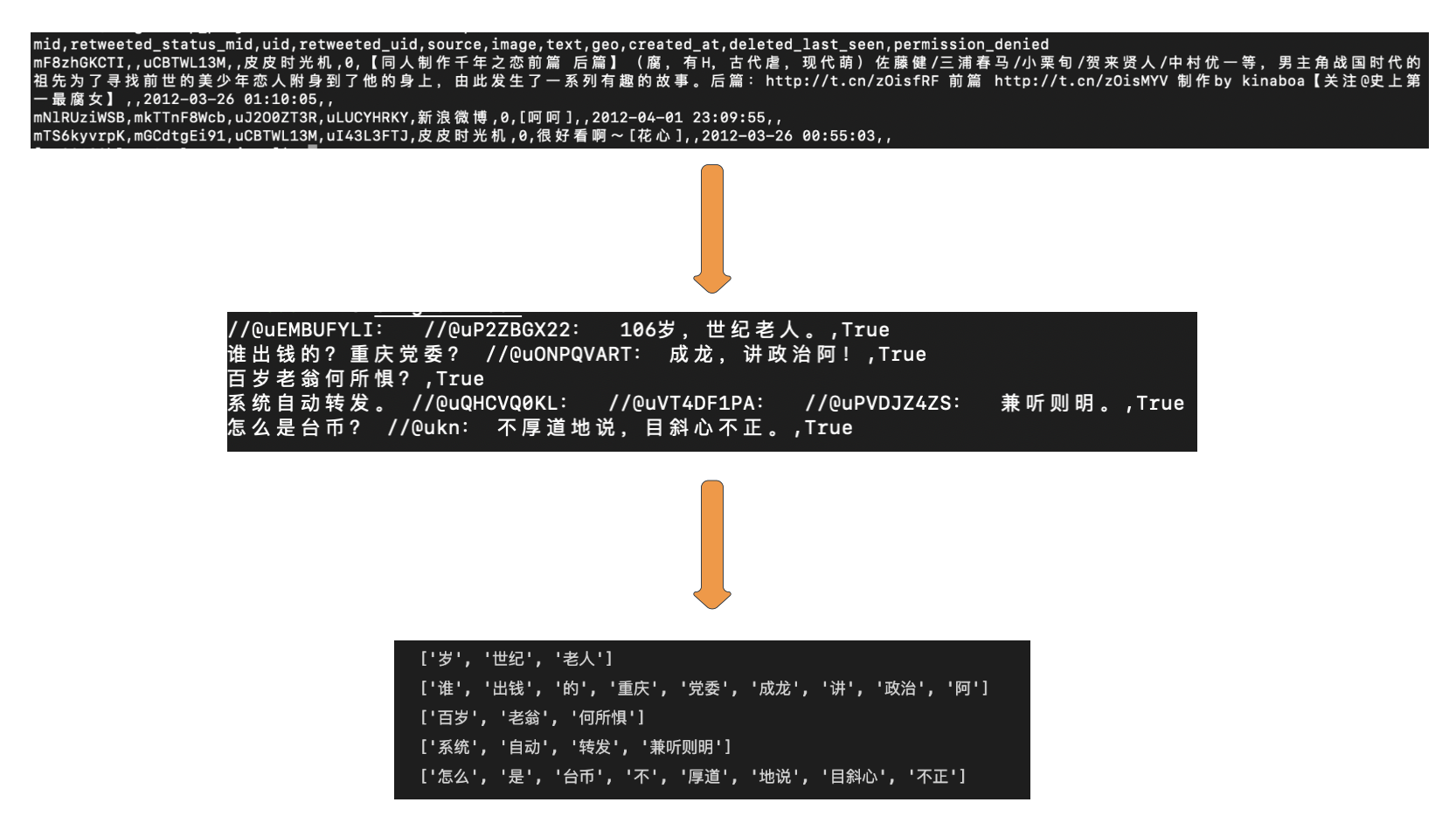}
\label{Fig. 2} 
\caption{}
\end{figure}
More precisely on the randomization strategy, over 60 zipped CSV files representing weeks of 2012 could be fetched from \citeauthor{Fu2021}. Through an automation process, we were able to extract all zipped files and append them into a merged CSV of around 19GB. After traversing through, we decided to randomly extract 1\% of the merged dataset, in which is still more than enough data. Since we are not looking into factors like user timelines or the user-id, this tactic may be applicable. As previous works suggest, the content of each blog and the included combined phrases is the primary indicator for censorship \cite{w2}.
\subsection{Test, Train, Development Split}
Following the common practices of NLP, we need to split the data into three ways, with the test corpus reserved for the evaluation. Since we have many sets of potential models to work with, as well as BERT, in which a transformer will further change the orientation of each phrase; it is logical to do the split on raw data. Furthermore, the proportion of the split is a randomized 60/20/20 split, with 60 being the training set. This allowed us to maintain the same training split across models, reducing inconsistencies.
\subsection{Jieba, Chinese tokenizer}
With the data represented under the blog-writing style of Weibo, unwanted tokens like punctuation and user-id will need to be removed. Therefore, we implemented \citeauthor{Jiebam}, a Chinese tokenization library to remove stopwords and such unwanted texts. Unless otherwise stated, Jieba is the tokenizer to clean each phrase before the modeling. In Figure 2, with the same queries as the second diagram, \citeauthor{Jiebam} combined with a regex searcher was able to extract meaningful phrases in Chinese. From names of individuals, city names, and adjectives, these word vectors are the basis for our modeling later on. Of course, rows without Chinese characters will be dropped. 
\subsection{Profiling}
After randomization and before the split, an overview of our dataset is the following: 274356 rows in total and 9755 marked as censored. With censored data being only 3\% of the sample size, we need to account for such imbalance during evaluation. Specifically, the macro-F1 score weighs for both of the classes in our classifier and the AUC score assesses different thresholds to reveal the accuracy of our models.
\section{Methodology \& Experiment}
\subsection{Data Analysis}
After cleaning the data, the word from each text block is tokenized and gathered into an array. The number of times which each word has appeared in the rich text is recorded along with the number of times that the word has appeared in a censored text, and put together with the word into two different dictionaries. Figure 3 lists the top 20 words that appeared most frequently in texts that are being censored. Column 1 is the word entry, column 2 is the English-translated version of the word, and column 3 is the times which the word appears in censored texts.
\begin{figure}[!htb]
\includegraphics[width=220px]{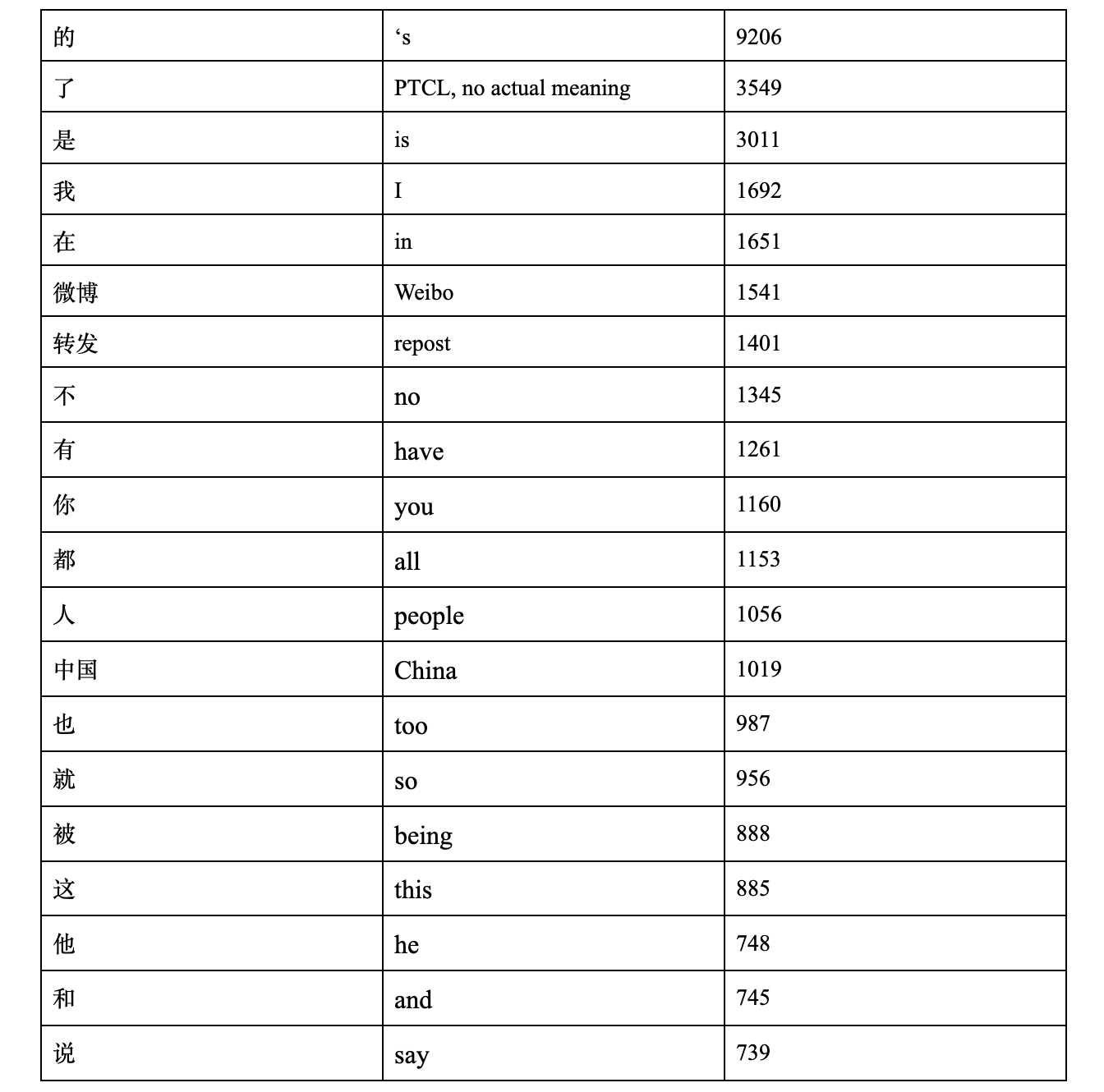}
\label{Fig. 3} 
\caption{}
\end{figure}

As the figure suggests, using the time which the word has appeared in censored texts is not a good practice. To gain useful information, the better practice is to divide the number of words appearing in censored texts with the number of total appearances of the word. Figure 4 lists the top 20 words that are most likely to be censored. Column 1 is the word entry, column 2 is the English translated version of the word, and column 3 is the probability of the entry being censored.

\begin{figure}[!htb]
\includegraphics[width=220px]{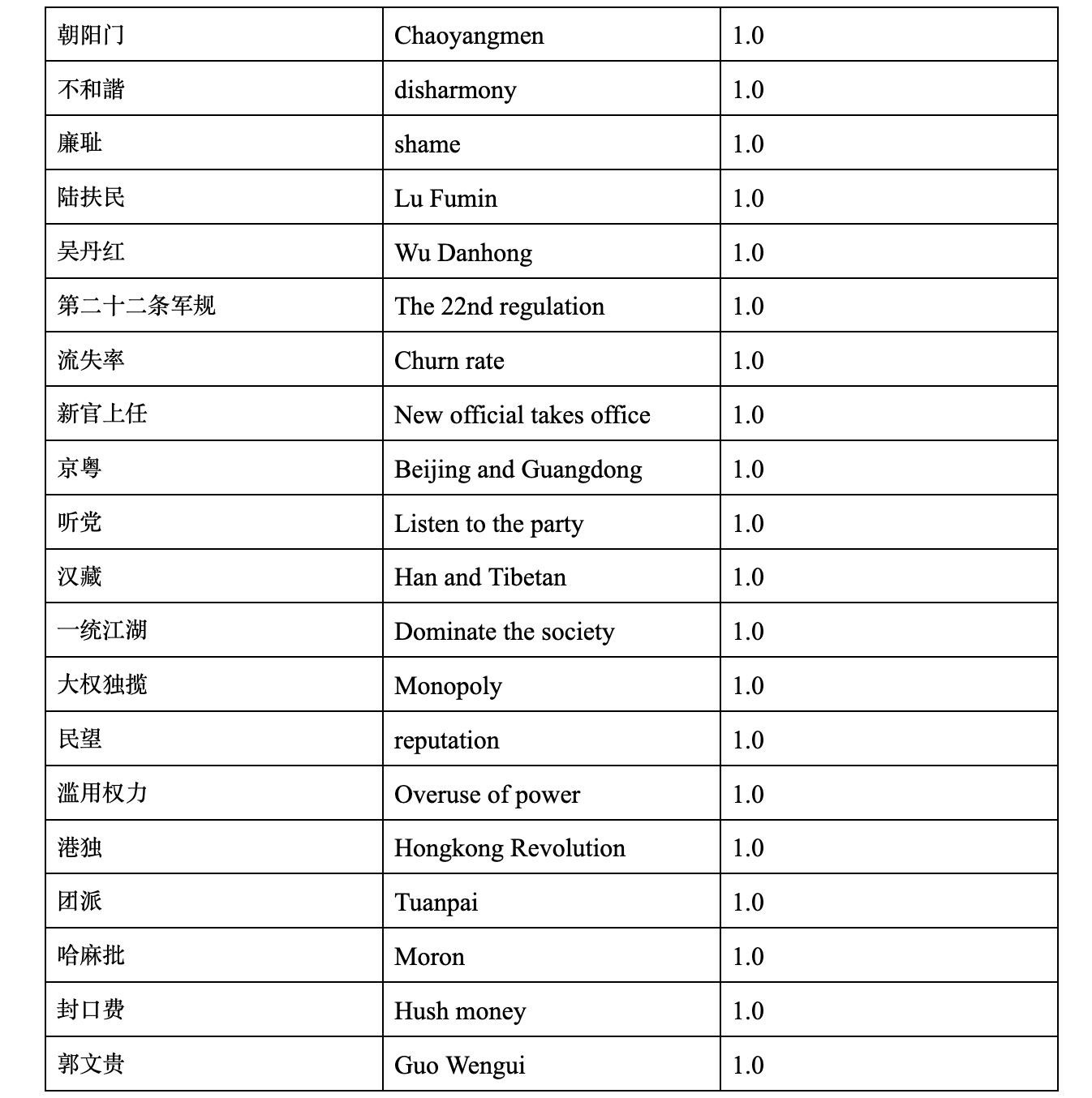}
\label{Fig. 4} 
\caption{}
\end{figure}

\subsection{Probability-based Models}

Using the dictionary of word entries and the probabilities of them being censored when appearing in a text, we developed four different logistic regression models and evaluated their performance. Each of these models takes one or more input variables that are calculated by the probability of words in a text. Using the discrete outcome of whether a text entry will be censored or not, based on the given dataset, four different logistic regression models have been generated.

\paragraph{Model 1: Single Most Probable Word}

This model uses only one input, which is the largest probability among the probabilities of words in a text. In this model, we only consider the word that is most likely to be censored in a text. This model is the simplest among four models, yet is more likely to be the censorship system of Weibo and other social media platforms in China \cite{m1}. A research conducted by indicated that in real life, if a user posts a text in social media platforms that contains words that the government considers to be insecure, which would fall into the category of sensitive word, the text will be temporarily held until administrators check it. This model simulates such process and uses the logistic regression algorithm to calculate the outcome of the text being censored instead of manual revision.

\paragraph{Model 2: Multinomial Probabilities}

This model is generated based on probabilities of all words appearing in the text. It considers all probabilities from words in a text as an array of floats with fixed length. We use 8 as the fixed array length. To resize each text entry to the fixed length, an array of the size of the number of words in a text is generated and filled with the probability of each word, then it is sorted based on the probability from most to least. Then the array is resized to the fixed length, and is used as the input variable of the logistic regression.

\paragraph{Model 3: Vector Length}

This model uses the similar idea as Model 1, however it considers the factors of other words inside the text. It considers the probabilities of each word in the text being censored as a vector. By inputting the length of the vector, the logistic regression model is generated. Compared to Model 1, Model 3 calculates the overall weight of the probability of a text being censored or not based on each of its word’s probability, and may hold advantage in some specific cases. For example, if we are comparing these two texts: “A B C D” and “A E F G”, with P(A) = 0.8, P(B) = 0.2, P(C) = 0.1, P(D) = 0.0, P(E) = 0.8, P(F) = 0.7, P(G) = 0.6, where P(X) represents the probability of word X being censored or not. The text “A E F G” appears to be more likely to be censored compared to “A B C D”, yet Model 1 will produce the same outcome. By using the vector length, Model 3 can fix such problem.

\paragraph{Model 4: Cosine Similarity}

This model is identical to the concept of TF-IDF score. The difference between them is that this model uses the probabilities of words, instead of the frequencies of words appearing in the document as TF-IDF suggests. The basic idea of this model is to think of a text that will be nearly 100\% likely to be censored, in other words, to construct an array filled with float 1.0 with the same size as the input text, then this model compares the similarity of the input text with the text that will be censored, and uses the cosine similarity to generate logistic regression model. 

\subsection{Transformers \& BERT}

\paragraph{Purpose:}
We attempt to apply the power of Transformer models to evaluate classification methods. One of the benefits of using Transformers compared to the previous approaches is the inclusion of positional embeddings in both inference and training. If we see a large discrepancy between the performance of the transformer and other models, we can infer that it is likely that the censorship algorithms are more complicated than bag-of-words approaches such as TF-IDF. Similarly, we also attempt to fine-tune a pre-trained BERT (Bidirectional Encoder Representations from Transformers) model to perform classification tasks for our dataset. \cite{bertpaper}

\includegraphics[width = 200pt]{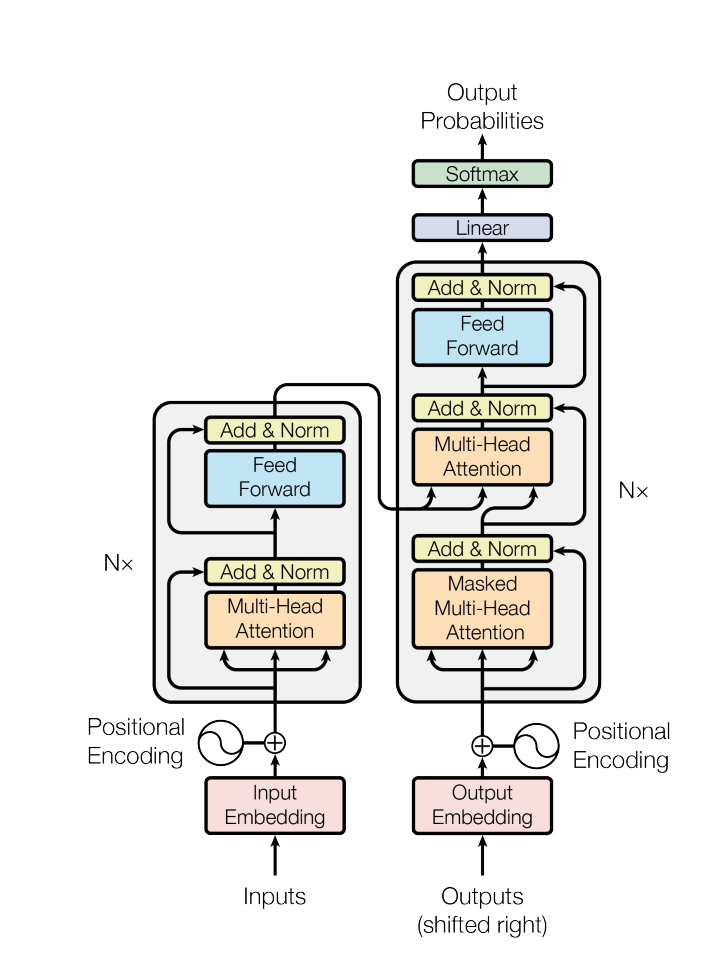}
\cite{DBLP:journals/corr/VaswaniSPUJGKP17}

\paragraph{Approach and Evaluation:}
The models are trained using class weights based on label frequencies to account for the data imbalanced. Additionally, binary cross-entropy is chosen as the loss metric, and the metric to minimize in all the below models. 
\paragraph{Model 5: Transformer from Scratch}

We build a model from scratch utilizing Keras \cite{chollet2015keras} based on the model architecture seen in Attention is All You Need by \citeauthor{DBLP:journals/corr/VaswaniSPUJGKP17}. Prior to being fed through the model, we generate vector embeddings by tokenizing sequences and converting tokens into indeces mapped by the tokenizer. These vectors are further padded/truncated to a maximal length of 200 tokens. The first layer of the model converts the token indeces into one-hot-encoded binary vectors, and positional embeddings are then concatenated to the vector using the following structure proposed by the authors of AIAYN:
\begin{align*}
    PE(pos, 2i) &= \sin{(pos/10000^{2i/d_{model}})}\\
    PE(pos, 2i+1) &= \cos{(pos/10000^{2i/d_{model}})}\\
\end{align*}
The vector embeddings then go through a single Transformer Block, Global Pooling, and Dense layers. Dropout is also applied during training to prevent the model from trying to overfit.
To perform tokenization, two methods were considered, one using \citeauthor{Jiebam}, and one using a pre-trained HuggingFace BERT-based tokenizer for Chinese. Although both methods indicated similar performance, the tokenizer in \textit{bert-base-chinese} was chosen \cite{huggingface} due to the helpful tools provided by the HuggingFace Transformers library. 

\paragraph{Model 6: Fine-Tuned BERT}
We attempt to fine-tune the \textit{bert-base-chinese} model provided by HuggingFace to perform classification tasks. With minimal modifications, we used the Transformers library to download the pre-trained model into PyTorch \cite{NEURIPS2019_9015} format. Since the model includes a trainable tokenizer, sentences are sent to the model without any need for preprocessing. We provided the model the classification task, and let it train for 3 epochs using the entire train set. Since even the base model is quite large, training took much longer than the previous model. 

\section{Results}
\subsection{Performance Evaluation}

\begin{table}[!htb]
\centering
\begin{tabular}{lll}
\hline \textbf{Model} & \textbf{Macro-F1} & \textbf{AUC} \\ \hline
Single Most Probable Word & 0.567 & 0.773 \\
Multinomial Probabilities & 0.596 & 0.781 \\
Vector Length & 0.586 & 0.823 \\
Cosine Similarity & 0.744 & 0.806 \\
Scratch Transformer & 0.754 & 0.893\\
Fine-Tuned BERT & 0.733 & 0.941\\
\hline
\end{tabular}
\caption{Model Performance }
\end{table}

Results are shown in Table 1, where Column 1 is the name of the model, column 2 is the macro-F1 score, and column 3 is the AUC score based on inference on the validation set.

\begin{figure}[!htb]
\includegraphics[width = 200pt]{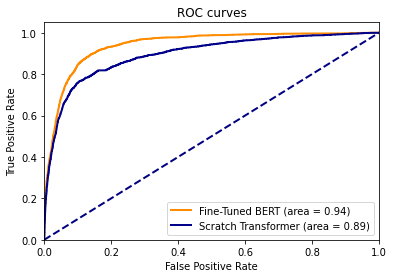}
\label{Fig. 3} 
\caption{}
\end{figure}

Figure 5 provides the visualization of the ROC curves for the two BERT-based models. This suggests that the BERT model is slightly better at capturing the tradeoff between True and False positive rates.

\subsection{Conclusions}

We considered two metrics to compare performance between the models. The AUC scores evaluate the theoretical performance of the models, and the macro-F1 scores evaluate the actual results. 

The traditional probability-based models have lower macro-F1 scores and AUC scores. The Single Most Probable Word model has the lowest overall score comparing to the others due to the nature of its simplicity. The Cosine Similarity Model has an overall good results among all four probability-based models, surpassing the model introduced by Shaikh Sahil Ahmed and Anand Kumar M. in 2021 which holds the score of 0.634. To improve the score, we introduce the latest research of BERT.
The best performing models in terms of F1 scores were the Transformer and Cosine Similarity model.  Our least informative model is as expected, the single most probable word model.
When considering AUC scores, which may be slightly more useful in considering the existence of impact of non-NLP variables, the Fine-Tuned Bert model outperforms the rest of the models.
After applying these methods, in the Transformer models, we notice slightly better ROC-AUC performance, but almost equal F1 measurements as the best Bag-of-Words model. One suspicion around why we have seen an F1-score upper bound is the existence of third variables which may impact censorship, such as time of publishing, geolocation data, and bot detection. Considering these third variables could be an interesting attempt to get closer to the true censorship algorithms used. In comparison, the performance of Model 6 compared to Model 5 shows a slight improvement in AUC score but for a negligible decrease in F1 score. To further consider is the size of the BERT based model, which required powerful GPU's and much more compute time than trying a Transformer, making it an impractical solution for the slight improvement it provides.

\section{Further Works}
One of the priorities for continuous study may be the incorporation of tools to account for this imbalanced data source. Even though it may be hard to achieve, due to the nature of censorship, we could also employ other methodologies during modeling, specifically random oversampling. Other measures of evaluation may be accounted for as well, apart from AUC and Macro F-1. One of the ideas that we had during the proposal stage was doing the test manually by incorporating Weibo bots to physically post the posts marked as censored under our prediction. This idea could also be plausible while working on future projects regarding the topic. Lastly, some previous works employed a full web scraper similar to Weiboscope, calling the Weibo open API to get in-date results. With this strategy, a more accurate dataset may be outputted considering how frequently the enforcement guidelines are from Weibo. Furthermore, a comparison between the 2012 models and 2022 models may be conducted to evaluate the tightening of censorship over this 10-year period. 

Regarding the use of Transformer-based models and LLMs (large language models), the performance seen in this paper is further evidence for the flexibility of these models. Because of their usability in a variety of scenarios, they provide one simple framework for almost any application. \textit{This is examined in great detail in "Attention is All You Need"\cite{DBLP:journals/corr/VaswaniSPUJGKP17}}. Clearly a drawback though is the cost of the models, both in terms of computational and space complexity. The custom Transformer model took only minutes to train, while the BERT model took hours with a high-compute GPU, whereas simpler models with similar explanatory power take orders of magnitude less computing power. Simpler models are thus quicker to iterate and modify, as well as more deterministic in how they can be fine-tuned, as opposed to the black-box system of Transformer-based models. 

\bibliography{anthology,custom}
\bibliographystyle{acl_natbib}

\end{document}